%% file: main.tex
\documentclass[10pt,twocolumn,letterpaper]{article}
\usepackage[pagenumbers]{cvpr}
\input{paper_preamble}
\makeatletter\AtBeginDocument{\begingroup\let\bibcite\@gobbletwo\externaldocument{supp}\endgroup}\makeatother

\def\confName{CVPR}
\def\confYear{2027}

\title{
Benchmarking Without Teaching: Unlearnable Robotic Demonstrations for Vision-Language-Action Models
}

\author{Jiaming He$^{1}$ ~~ Shizhe Zhang$^1$ ~~ Jingsong Liang$^1$ ~~Yi Yu$^{2}$\thanks{Corresponding author.} ~~ Chen Lv$^1$ ~~ Wanhua Li$^1$ ~~ Xudong Jiang$^1$
\\
$^1$Nanyang Technological University ~\quad
$^2$Jilin University
\\
{\tt\small \{jhe040, exdjiang\}@ntu.edu.sg ~\quad yuyi26@jlu.edu.cn}
}

\begin{document}
\maketitle
\input{sec/0_abstract}
\input{sec/1_intro}
\input{sec/2_related}
\input{sec/3_method}
\input{sec/4_experiments}
\input{sec/5_conclusion}

\newpage
\input{sec/6_llm_usage}
{
  \small
  \bibliographystyle{ieeenat_fullname}
  \bibliography{references}
}

\clearpage
\newpage
\appendix
\renewcommand{\thesection}{S\arabic{section}}
\appendix
\renewcommand{\thesection}{S\arabic{section}}
\setcounter{figure}{0}
\renewcommand{\thefigure}{S\arabic{figure}}
\setcounter{table}{0}
\renewcommand{\thetable}{S\arabic{table}}
\setcounter{theorem}{0}
\renewcommand{\thetheorem}{S\arabic{theorem}}
\setcounter{proposition}{0}
\renewcommand{\theproposition}{S\arabic{proposition}}
\input{sec/X_suppl}

\end{document}

%% file: paper_preamble.tex
\input{cvpr_preamble}
\usepackage{iftex}
\ifXeTeX
  \usepackage{fontspec}
\fi
\usepackage{amsmath,amssymb,amsthm}
\input{preamble}
\renewcommand{\method}{\textsc{Mute}\xspace}  
\usepackage{enumitem}
\usepackage{algorithm,algpseudocode}

\newtheorem{proposition}{Proposition}
\theoremstyle{definition}

\usepackage{xr-hyper}  
\definecolor{cvprblue}{rgb}{0.21,0.49,0.74}
\usepackage[pagebackref,breaklinks,colorlinks,allcolors=cvprblue]{hyperref}
\makeatletter\AtBeginDocument{\def\hyper@linkfile#1#2#3{#1}}\makeatother

%% file: cvpr_preamble.tex
\usepackage{cuted} 

\usepackage{currfile} 

\usepackage{caption} 



%% file: preamble.tex
\usepackage{booktabs}
\usepackage{multirow}
\usepackage{colortbl}      
\usepackage{makecell}
\usepackage{pifont}        
\usepackage[most]{tcolorbox}
\usepackage{xspace}

\definecolor{cblue}{HTML}{2F6FDE}
\definecolor{cteal}{HTML}{0E9AA7}
\definecolor{cgreen}{HTML}{2BA84A}
\definecolor{camber}{HTML}{F2A541}
\definecolor{corange}{HTML}{EE7733}
\definecolor{cred}{HTML}{E0464E}
\definecolor{cpink}{HTML}{D9418C}
\definecolor{cpurple}{HTML}{7B5CD6}
\definecolor{cindigo}{HTML}{3B4BA8}
\definecolor{cink}{HTML}{1D2433}
\definecolor{cmuted}{HTML}{8A93A6}

\definecolor{tabhead}{HTML}{E5E5E5}   
\definecolor{tabzebra}{HTML}{F2F2F2}  
\definecolor{tabours}{HTML}{D7F6FF}   
\definecolor{gaincolor}{HTML}{1E8E3E}
\definecolor{losscolor}{HTML}{D93025}

\usepackage{adjustbox}

\newcommand{\method}{\textsc{Name}\xspace}

\newtcolorbox{rqbox}{
  enhanced, breakable=false, colback=white, colframe=black!65, boxrule=0.6pt, arc=7pt, outer arc=7pt,
  left=8pt, right=8pt, top=3pt, bottom=3pt, before skip=6pt, after skip=6pt, fontupper=\itshape}

\definecolor{thmframe}{HTML}{4E7F73}
\definecolor{thmback}{HTML}{F5FAF8}
\newtcolorbox{thmbox}{enhanced, breakable=false, colback=thmback, colframe=thmframe, boxrule=0.9pt, arc=4pt,
  outer arc=4pt, left=6pt, right=6pt, top=3pt, bottom=3pt, before skip=6pt, after skip=6pt}

%% file: sec/0_abstract.tex
\begin{abstract}

A public benchmark for offline policy evaluation also exposes the same robot demonstrations to unauthorized training.
Existing unlearnable examples can undermine training, but preserving the appearance of demonstrations does not ensure that frozen policies make comparable predictions on them.
We introduce \method to protect robot demonstrations for vision-language-action models while preserving their value as an offline benchmark.
The method builds on an asymmetry between training and evaluation: a perturbation can be a shortcut for a learner even when it barely changes a frozen policy's predictions.
\method encodes continuous action windows as weights over shared noise patterns, selecting patterns that strongly activate pretrained visual encoders while eliciting weak responses from frozen reference policies.
It places these patterns on the least sensitive image tiles and verifies both action-error changes and prediction drift before release.
Across three LIBERO suites, SmolVLA trained on protected demonstrations achieves at most 7\% task success, compared with 74--85\% of clean training, while the reference's action error changes by at most 2\%.
Protection also extends to Meta-World, RoboTwin and SO-100, and transfers to OpenVLA.
On a real bimanual robot, $\pi_0$ trained on protected demonstrations fails every trial for which clean training achieves high success.
\end{abstract}

%% file: sec/1_intro.tex
\section{Introduction}
\label{sec:intro}

\begin{figure}[t]
  \centering
  \includegraphics[width=\columnwidth]{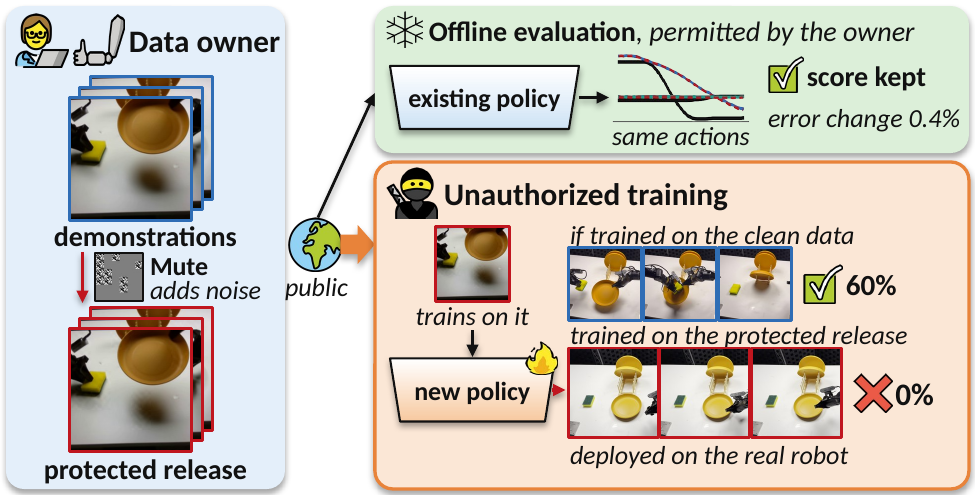}
  \vspace{-5mm}
  \caption{\textbf{Unlearnable demonstrations for robot learning.}
A data owner publishes a protected release by perturbing only the images within each observation, leaving robot states, language instructions and action labels unchanged.
Existing policies can still be evaluated on it, while a new policy trained on it fails in the clean environment.
Frames, rollouts and numbers come from $\pi_0$ on the real dish-washing task at $\epsilon=\frac{4}{255}$.}
  \label{fig:teaser}
  \vspace{-2mm}
\end{figure}


Vision-language-action models (VLAs) translate visual observations and language instructions into robot actions.
Recent models such as RT-2~\cite{brohan2023rt2}, OpenVLA~\cite{kim2024openvla} and $\pi_0$~\cite{black2024pi0} rely on substantial collections of robot demonstrations.
Collecting these demonstrations requires considerable effort, motivating shared resources such as Open X-Embodiment~\cite{oneill2024open} and LIBERO~\cite{liu2023libero}.
However, public access also exposes these demonstrations to \textit{unauthorized training}.
As shown in \cref{fig:teaser}, a data owner may wish to release demonstrations as an \textit{evaluation-only benchmark}: researchers can compare existing policies by measuring their action predictions against the demonstrated actions, without using these observation-action pairs for training.
Such a benchmark supports reproducible policy comparisons, maintains the separation between evaluation and training data, and preserves the owner's control over training use.
The central challenge is to prevent unauthorized policy training on released demonstrations while preserving their utility as an offline evaluation benchmark.


Unlearnable examples (UEs)~\cite{huang2021unlearnable} prevent unauthorized training by adding small, invisible perturbations to released data.
Many UE methods introduce label-predictive features that encourage models to rely on perturbations rather than image content~\cite{yu2022availability}.
This principle supports data protection for classification~\cite{fu2022robust,sadasivan2023cuda}, segmentation~\cite{sun2024unseg} and multi-task dense prediction~\cite{yu2025mtlue}.
While these methods preserve visual appearance, their objectives do not explicitly constrain frozen-model predictions on protected inputs.


Preserving visual appearance alone is insufficient for an evaluation-only benchmark: even small perturbations can alter a frozen policy's action predictions and distort its evaluation scores.
Moreover, robot demonstrations provide continuous action labels, often as windows of future actions, making discrete class-wise shortcuts difficult to apply directly.
Effective protection must therefore induce shortcuts for continuous action prediction while limiting changes to frozen-policy predictions and evaluation scores.
This raises the following question:
\emph{Can released robot demonstrations remain a faithful offline benchmark for existing policies while discouraging unauthorized policy training?}

Our empirical analysis reveals that a perturbation can provide a predictive training shortcut while having little effect on a frozen policy's outputs.
On LIBERO-Spatial, action-conditioned noise causes much less evaluation distortion than random noise at the same budget, yet a policy trained on it fits the protected demonstrations and fails in the clean environment.
Our theoretical analysis explains this separation from two complementary perspectives.
A least-squares model shows how reliable action-predictive perturbations can dominate learning and increase error on clean inputs, while a local Jacobian analysis characterizes their effect on frozen-policy predictions.
We further observe that reference-policy sensitivity is concentrated in a few image regions, whereas pretrained encoders respond more evenly.
Together, these findings motivate shortcuts that undermine policy learning while preserving policy evaluation.

Based on these findings, we propose \method to induce training shortcuts while preserving offline evaluation.
To encode continuous action windows, a compact trajectory embedding retains the current action and summarizes future motion through residual components, whose coordinates weight shared noise patterns.
Guided by the local Jacobian analysis, null-response patterns are selected to elicit strong pretrained-encoder responses while limiting reference-policy responses, making the action signal accessible to learners with little effect on frozen-policy predictions.
An adaptive support further exploits spatial sensitivity differences by placing noise on the least responsive image tiles.
Since this analysis is local, a final check verifies action-error changes and prediction drift on the saved images before release.
Experiments across simulated benchmarks, real-world demonstrations and policy architectures demonstrate strong protection with limited evaluation distortion.
On three LIBERO suites, SmolVLA success falls to at most 7.0\% while both evaluation budgets are met. On a real bimanual robot, protected training yields zero success on two tasks where clean training reaches 90\% and 60\%.


Our contributions are summarized as follows:
\begin{itemize}[leftmargin=1.2em,itemsep=1pt,topsep=2pt]
  \item We are the first to formulate robot demonstration protection that prevents unauthorized policy training while preserving offline evaluation utility, using image-only perturbations under explicit bounds on evaluation error changes and prediction drift.
  \item Empirical observations and theoretical analysis motivate \method, which encodes continuous action windows into training shortcuts and jointly selects noise patterns and spatial support to favor encoder responses while suppressing frozen-policy responses.
  \item Extensive experiments on three simulated benchmarks and three policy architectures show strong protection with limited evaluation distortion, supported by transfer and robustness analyses and real-robot trials yielding zero success on two tasks after protected training.
\end{itemize}

%% file: sec/2_related.tex
\section{Related Work and Motivation}
\label{sec:related}

\textbf{Unlearnable Examples.}
Huang et al.~\cite{huang2021unlearnable} introduced error-minimizing noise to prevent models from learning useful representations from protected images.
Subsequent work linked this effect to shortcut learning~\cite{geirhos2020shortcut,yu2022availability}, and developed alternative constructions based on adversarial perturbations~\cite{fowl2021adversarial,wang2026dualbranch}, autoregressive processes~\cite{sandoval2022autoregressive} and one-pixel patterns~\cite{wu2023onepixel}.
Universal generators protect many concepts~\cite{chen2024oneforall} and segmentation as in UnSeg~\cite{sun2024unseg}, and other extensions protect point clouds~\cite{wang2024unlearnable3d} and event streams~\cite{wang2025uevs}.
Related perturbations guard images and voices against generative misuse~\cite{zhang2026identityguard,xiong2026noway,qu2024dfrap,hu2026voicecloak,song2024geometrycloak}.
Protection remains dependent on the training procedure, since pretrained weights can weaken it~\cite{li2026bait,wang2026channel}.
Counter-protection methods rely on compression~\cite{liu2023iss}, augmentation~\cite{qin2023ueraser,dang2026adversarial,li2025detecting} and diffusion purification~\cite{dolatabadi2024avatar,wang2024eclipse}.

\vspace{1mm}\noindent\textbf{Data Protection in Robot Learning.}
Test-time attacks optimize visual perturbations to change the actions of VLAs~\cite{wang2025vlaadversarial} and VLM-based embodied agents~\cite{wang2025advedm}.
VLA backdoors use image triggers~\cite{zhou2025badvla,xu2025dropvla}, instruction words~\cite{buhler2026imperio,liu2026trapvla} or modified robot states~\cite{guo2026state} to induce a selected behavior.
Backdoors and jailbreaks also compromise VLM-based manipulation~\cite{wang2024robotcollapse} and embodied LLM agents~\cite{zhang2025badrobot}.
Kalra et al.~\cite{kalra2025dataset} modify demonstration images while retaining expert action labels to influence a behavior-cloning policy.
Other work acts on a trained model, as GuardVLA~\cite{sun2026guardvla} verifies ownership through a watermark and demonstration unlearning~\cite{li2026demonstration} edits a policy to remove selected demonstrations.
Contamination-resistant benchmarks~\cite{allawati2026contamination} and anti-distillation sampling~\cite{savani2025antidistillation} consider a related separation between inference utility and learning from released text.

\vspace{1mm}\noindent\textbf{Shortcuts in Imitation Learning.}
Behavior cloning can accumulate errors as a policy departs from the states represented in its demonstrations~\cite{ross2010efficient,ross2011reduction}.
It is also vulnerable to causal confusion~\cite{dehaan2019causal} and reliance on past actions that predict the next training label~\cite{wen2020fighting}.
Accurate prediction on demonstrations therefore need not imply successful behavior, and gradient starvation~\cite{pezeshki2021gradient} explains how a readily learned feature can suppress learning from other features.

\vspace{1mm}\noindent\textbf{Motivation of Our Work.}
These studies motivate protecting robot demonstrations against unauthorized policy training while preserving offline evaluation utility.
This requires training shortcuts suited to continuous action windows with limited effects on frozen policies.
Our approach combines action-conditioned noise with patterns and spatial support selected through policy sensitivity to limit prediction drift and evaluation error changes.
Only images are perturbed, with all other demonstration components unchanged.

%% file: sec/3_method.tex
\begin{figure*}[t]
  \centering
  \includegraphics[width=\textwidth]{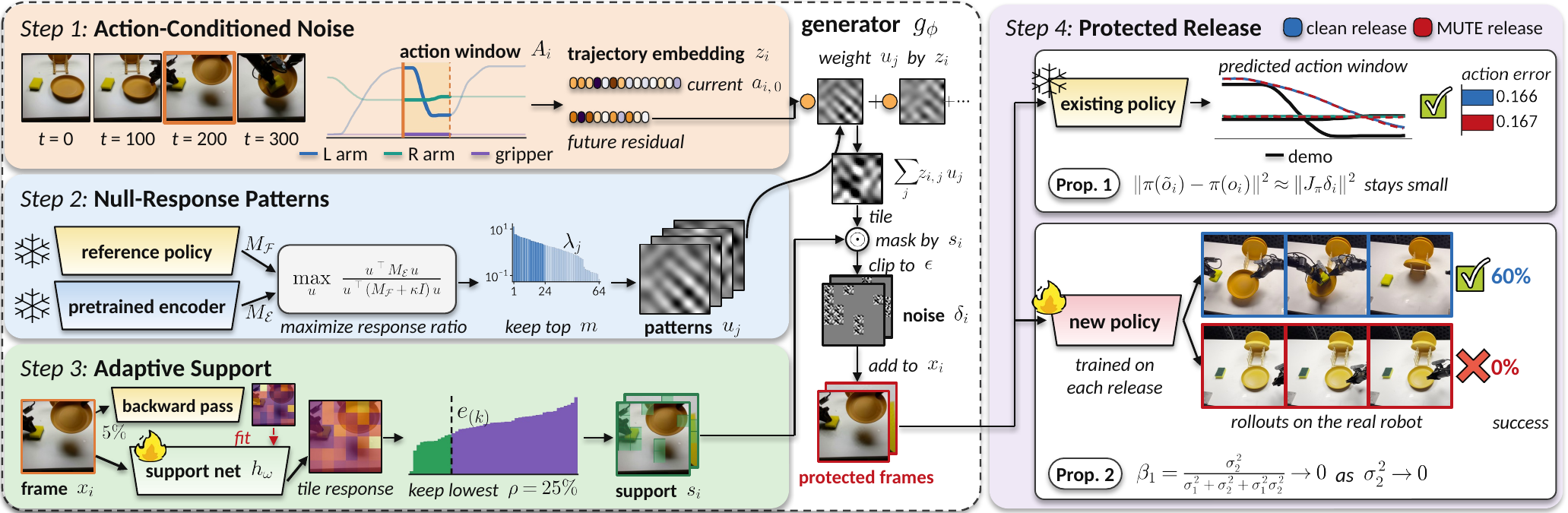}
  \vspace{-6mm}
  \caption{\textbf{Overview of \method.} The dashed frame is the generator $g_\phi$, drawn with its real intermediates on one real dish-washing frame under the settings of the real-robot release. The trajectory embedding $z_i$ weights the null-response patterns $u_j$, and their sum is tiled over the support $s_i$ predicted by the support network $h_\omega$. On the right, a frozen policy predicts the same actions on the protected frame, and a policy trained on the release fails on the real robot.}
  \label{fig:framework}
  \vspace{-2mm}
\end{figure*}

\section{Method}
\label{sec:method}

\textbf{Overview.} Our goal is to \textbf{1)} reduce the training value of released demonstrations while \textbf{2)} preserving their use for offline policy evaluation.
\Cref{fig:framework} summarizes the resulting procedure:
(1) A trajectory embedding of each action window sets the coefficients of action-conditioned noise.
(2) Null-response patterns produce a strong response in pretrained encoders and a weak response in reference policies.
(3) An adaptive support places the noise on the image tiles to which the reference policies respond least.
(4) A utility check verifies evaluation fidelity before release.
The first three steps form the generator $g_\phi$, which maps a frame and its action window to a perturbation in one forward pass.

\subsection{Problem Formulation}
\label{sec:formulation}

\noindent \textbf{Demonstrations and policies.}
A demonstration set $\mathcal D=\{(o_i,A_i)\}_{i=1}^{n}$ contains $n$ frames.
The observation $o_i=(x_i,q_i,\ell_i)$ consists of a camera image $x_i$ with pixel values in $[0,1]$, the robot state $q_i$ and the language instruction $\ell_i$.
Its label is the action window $A_i=[a_{i,0},\ldots,a_{i,H-1}]^{\top}\in\mathbb R^{H\times d}$, where $a_{i,k}\in\mathbb R^{d}$ is the action taken $k$ steps after frame $i$ and $H$ is the prediction horizon.
Actions are standardized with the mean and standard deviation of the dataset.
A policy $\pi_\theta$ with parameters $\theta$ maps an observation to a predicted window $\pi_\theta(o)\in\mathbb R^{H\times d}$.
When a robot has several cameras, each view is processed in the same way with its own patterns, so we describe a single view.

\vspace{1mm} \noindent \textbf{Protected release.}
The owner perturbs the images only, and keeps the states, instructions and actions unchanged.
The protected release $\widetilde{\mathcal D}=\{(\tilde o_i,A_i)\}_{i=1}^{n}$ contains
\begin{equation}\small
\label{eq:dataset}
\tilde o_i\!=\!(\operatorname{clip}_{[0,1]}(x_i\!+\!\delta_i),\,q_i,\,\ell_i),~
\delta_i\!=\!g_\phi(x_i,A_i),~
\|\delta_i\|_\infty\!\le\!\epsilon ,
\end{equation}
where $\operatorname{clip}_{[0,1]}$ is clipping to the valid range, $g_\phi$ is the generator with parameters $\phi$, and $\epsilon$ is the perturbation budget.

\vspace{1mm} \noindent \textbf{Training on the release.}
A downstream user trains a policy on the release by minimizing its behavior-cloning loss:
\begin{equation}\small
\label{eq:bc}
\theta^\star=\arg\min_{\theta}\ \frac1n\sum_{i=1}^{n}\mathcal L(\theta;\tilde o_i,A_i),
\end{equation}
where $\mathcal L$ is the sample loss of the model and $\theta^\star$ the trained parameters.
Details of the flow-matching and token losses of the policies are given in Appendix~\ref{sec:training_losses}.

\vspace{1mm} \noindent \textbf{Evaluation utility.}
The protected release must also serve the offline evaluation of a frozen reference policy $\pi$.
We measure the action error of $\pi$ and its prediction drift:
\begin{equation}\small
\label{eq:offline}
\begin{aligned}
E_\pi(\mathcal D)\!=\!\frac{\sum_{i=1}^{n}\!\big\|\pi(o_i)\!-\!A_i\big\|^2}{nHd},~\Delta_\pi\!=\!\frac{\sum_{i=1}^{n}\!\big\|\pi(\tilde o_i)\!-\!\pi(o_i)\big\|^2}{nHd}.
\end{aligned}
\end{equation}
Here $\|\cdot\|$ is the Euclidean norm of the flattened matrix, and the protected error $E_\pi(\widetilde{\mathcal D})$ replaces $o_i$ with $\tilde o_i$.
For a policy that samples its actions, both predictions in $\Delta_\pi$ share the same sampling noise.
The error reflects the evaluation score, and the drift measures changes in individual predictions that may cancel in that score.

\vspace{1mm} \noindent \textbf{Protection objective.}
Let $S(\theta)$ be the success rate of $\pi_\theta$ in the clean environment, and let $\mathcal F$ be a set of frozen reference policies.
The owner seeks the generator $g_\phi$ that solves
\begin{equation}\small
\label{eq:objective}
\begin{aligned}
&\min_{g_\phi}
\underbrace{S(\theta^\star)}_{\text{Goal (1): reduce training value}}\\
&\text{s.t.}\quad
\|\delta_i\|_\infty\le\epsilon,\quad i=1,\ldots,n,\\
&\underbrace{\Delta E\!=\!\frac{\big|E_\pi(\widetilde{\mathcal D})\!-\!E_\pi(\mathcal D)\big|}{E_\pi(\mathcal D)}
\!\le\!\tau_E,~\textbf{Drift}\!=\!\frac{\Delta_\pi}{E_\pi(\mathcal D)}\!\le\!\tau_D,~~\pi\!\in\!\mathcal F}_{\text{Goal (2): preserve invisibility and offline evaluation}}.
\end{aligned}
\end{equation}
Here $\theta^\star$ is the solution of \cref{eq:bc} and depends on $g_\phi$ through the release.
The tolerances $\tau_E$ and $\tau_D$ bound the relative change of the action error and the relative drift.
Both constraints are needed because a small change of the error can hide a large drift (as in Appendix \ref{sec:utility_bound}).
We use $\epsilon=\frac{4}{255}$, $\tau_E=2\%$ and $\tau_D=6\%$ unless stated otherwise.

\vspace{1mm} \noindent \textbf{Threat model.}
Following the standard threat model of UEs~\cite{huang2021unlearnable,fu2022robust,sun2024unseg,yu2025mtlue}, the owner knows the demonstrations but not the downstream architecture or training recipe, and can use public reference policies $\mathcal F$ and pretrained image encoders $\mathcal E$.
The user may choose any training procedure but has no additional demonstrations.
Since the learner is unknown, we build an action-conditioned shortcut and choose its patterns and support with tractable response objectives instead of differentiating through \cref{eq:bc}.

\begin{figure}[t]
  \centering
  \includegraphics[width=\columnwidth]{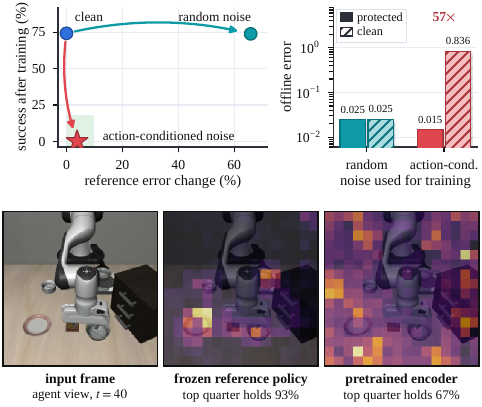}
  \vspace{-7mm}
  \caption{\textbf{Noise effects on evaluation and training at $\epsilon\!\!=\!\!\frac{8}{255}$.} Upper left: reference error change and success of SmolVLA trained on each release (target region in green). Upper right: offline error of the trained policy. Lower row: an input frame and its first-order tile responses, where brighter tiles respond more strongly.
  Action-conditioned noise is our method to be introduced in Sec.~\ref{sec:noise}.}
  \label{fig:observation}
  \vspace{-4mm}
\end{figure}

\subsection{Empirical Observation}
\label{sec:observation}

We compare noise constructions on LIBERO-Spatial~\cite{liu2023libero} at $\epsilon=\frac{8}{255}$.
For each release, we fine-tune SmolVLA~\cite{shukor2025smolvla} and evaluate it for 500 episodes in the clean simulator.
A public SmolVLA policy trained on clean LIBERO serves as the frozen reference. The result is presented in \cref{fig:observation}.

\vspace{1mm} \noindent \textbf{Noise magnitude doesn't determine its effect on training.}
The upper-left plot of \cref{fig:observation} places each release by its evaluation distortion and by the success of policies trained on it.
Random sign noise raises the action error of the frozen reference by 66\%.
Nevertheless, training on this release achieves 73.8\% success compared with 74.2\% for clean training.
Action-conditioned noise changes the reference error by 3.8\% while reducing the success of the trained policy to 0.0\%.
Thus, perturbations at the same pixel budget can have very different effects on evaluation and training.

\vspace{1mm} \noindent \textbf{Accurate prediction on protected frames can hide shortcut dependence.}
The upper-right plot examines what the policy learns from action-conditioned noise.
Its offline error is 0.015 on protected frames and 0.84 on clean frames.
This gap indicates strong dependence on the perturbation.
The pattern used to encode the action information affects both outcomes.
Per-patch color offsets increase the error of the frozen reference by 27\%, while random tile patterns leave the trained policy with 8.8\% success.
These comparisons motivate selecting the noise patterns explicitly.

\noindent \textbf{Spatial response provides a second design choice.}
The lower row compares the first-order response of each $16\times16$ tile for the frozen reference and the pretrained encoder on one frame.
The most responsive quarter of the tiles carries 93\% of the frozen-policy response and 67\% of the encoder response.
On the least responsive half, the reference responds per tile with 4\% of its response under uniform tiling, while the encoder keeps 103\% of its response per tile.
These measurements motivate adapting the support to each image.

The core observation is \emph{a perturbation can be predictive during training while causing little change in a frozen policy's outputs}.
Its correlation with the actions motivates the noise coefficients, and the responses of reference policies and encoders motivate the patterns and their support.

\subsection{Theoretical Analysis}
\label{sec:theory}

We separate the information that a perturbation supplies during training from the local response of a frozen policy.
For training, a least-squares model shows that a feature predicting the action more reliably than the image content receives a larger weight, so the error on clean inputs grows with its reliability (Proposition S\ref{prop:weight} in Appendix~\ref{sec:proof_weight}).


\vspace{1mm} \noindent \textbf{Response of a frozen policy.}
We divide the image into $P$ disjoint tiles of $b\times b$ pixels.
Let $T_p$ be the linear map that writes a tile pattern $u\!\in\!\mathbb R^{3b^2}$ into tile $p$, setting all other pixels to zero.
A perturbation repeating $u$ on tiles selected by a binary support $s\!\in\!\{0,1\}^{P}$ is
$\delta\!=\!\sum_{p=1}^{P}s_pT_pu$.
Let $J_\pi$ be the Jacobian of the flattened prediction $\pi(o)$ with respect to the image $x$, with other inputs and sampling noises fixed.
\begin{proposition}[Local prediction response]
\label{prop:inert}
Suppose $\pi$ has a bounded second derivative near $x$, and let $k=\sum_{p}s_p$ be the number of selected tiles.
Then
\vspace{-2mm}
\begin{equation}\small
\label{eq:quadform}
\begin{aligned}
\|\pi(\tilde o)-\pi(o)\|^2
&=\Big\|\sum_{p=1}^{P}s_p\,J_\pi T_pu\Big\|^2+O\big(\|\delta\|^3\big)\\[-4pt]
&\le k\sum_{p=1}^{P}s_p\,\|J_\pi T_pu\|^2+O\big(\|\delta\|^3\big).
\end{aligned}
\end{equation}
\end{proposition}
\vspace{-1mm}

Appendix~\ref{sec:proof_inert} gives the proof.
The first line is a quadratic form in $u$, which \method uses to choose pattern directions with a small response.
The second line adds the responses of single tiles, which \method uses to choose the support tile by tile.
Both forms are local, so the saved images are checked with the full policies.

\subsection{Action-Conditioned Noise}
\label{sec:noise}

The first step turns each action window into the coefficients of a learning shortcut.
While class-wise UEs encode discrete labels~\cite{huang2021unlearnable,yu2022availability}, we encode continuous action windows into compact vectors whose entries weight shared patterns.

\vspace{1mm} \noindent \textbf{Trajectory embedding.}
The embedding keeps the current action $a_{i,0}$ exactly and summarizes the future actions $y_i=[a_{i,1};\ldots;a_{i,H-1}]\in\mathbb R^{(H-1)d}$.
Future actions are strongly correlated with the current one, so we first remove their linear dependence on it,
\vspace{-2mm}
\begin{equation}\small
\label{eq:residual}
\begin{aligned}
(W^\star,c^\star)\!=\!\arg\!\min_{W,c}\sum_{i=1}^{n}\!\big\|y_i\!-\!Wa_{i,0}\!-\!c\big\|^{\!2}\!,~r_i\!=\!y_i\!-\!W^\star a_{i,0}\!-\!c^\star\!,
\end{aligned}
\end{equation}
where $W$ is a regression matrix and $c$ an intercept, and the residual $r_i$ is the part of the future motion that the current action does not predict.
Let $B\in\mathbb R^{(H-1)d\times K}$ hold the $K$ leading eigenvectors of the residual covariance $\Sigma_r=n^{-1}\sum_{i}r_ir_i^\top$ and $\Lambda\in\mathbb R^{K\times K}$ the diagonal matrix of their eigenvalues.
The trajectory embedding is
\vspace{-2mm}
\begin{equation}\small
\label{eq:embedding}
z_i=\begin{bmatrix}a_{i,0}\\[1pt]\Lambda^{-1/2}B^\top r_i\end{bmatrix}\in\mathbb R^{m},\qquad m=d+K .
\end{equation}
For example, a command that closes the gripper stays in $z_i$ even when it explains little variance of the whole window.

\vspace{1mm} \noindent \textbf{Noise synthesis.}
Let $U=[u_1,\ldots,u_m]\in\mathbb R^{3b^2\times m}$ hold one tile pattern per embedding coordinate, and let $s_i\in\{0,1\}^{P}$ be the support of frame $i$.
The generator writes the weighted sum of the patterns into every selected tile,
\vspace{-2mm}
\begin{equation}\small
\label{eq:write}
\delta_i=\Pi_\epsilon\Big(\frac{\gamma\epsilon}{\sqrt m}\sum_{p=1}^{P}s_{i,p}\,T_pUz_i\Big),
\end{equation}
where $\gamma$ is a gain and $\frac{1}{\sqrt m}$ keeps the tile signal $Uz_i$ at a similar scale for different embedding sizes.
$\Pi_\epsilon$ clips each value to $[-\epsilon,\epsilon]$, keeps $x_i+\delta_i$ in $[0,1]$ and rounds it to 8 bits.
Before clipping, the perturbation is linear in $z_i$, so frames with similar action windows receive similar noise.

\subsection{Null-Response Patterns}
\label{sec:patterns}

The second step chooses patterns to which pretrained encoders respond strongly and the reference policies respond weakly.
The encoder response makes the noise easy to learn, and the weak policy response preserves evaluation.

\vspace{1mm} \noindent \textbf{Response matrices.}
Each pattern is restricted to low spatial frequencies by $\Phi\in\mathbb R^{3b^2\times3L^2}$, containing orthonormal two-dimensional discrete cosine atoms with the $L$ lowest frequencies along each axis of each color channel.
For a model $f$ in $\mathcal F$ or $\mathcal E$, let $J_f(x_i)$ be its output Jacobian with respect to the image, where the output is a policy's predicted window or an encoder's image embedding.
Under uniform tiling $T=\sum_{p=1}^{P}T_p$, the first line of \cref{eq:quadform} gives the response $v^\top\Phi^\top G_f\Phi v$ of model $f$ to the pattern $\Phi v$ with
\vspace{-2mm}
\begin{equation}\small
\label{eq:response}
G_f=\frac1n\sum_{i=1}^{n}T^\top J_f(x_i)^\top J_f(x_i)\,T .
\end{equation}
We estimate it on a random subset of frames with random output projections (Appendix~\ref{sec:spectral_solution}).
We normalize each model by its trace and average within each group,
\begin{equation}\small
\label{eq:response_ensemble}
\begin{aligned}
M_{\mathcal F}=\frac1{|\mathcal F|}\sum_{\pi\in\mathcal F}\frac{\Phi^\top G_\pi\Phi}{\operatorname{tr}(G_\pi)},~M_{\mathcal E}=\frac1{|\mathcal E|}\sum_{f\in\mathcal E}\frac{\Phi^\top G_f\Phi}{\operatorname{tr}(G_f)} .
\end{aligned}
\end{equation}

\vspace{1mm} \noindent \textbf{Pattern optimization.}
After scaling $M_{\mathcal F}$ to unit mean eigenvalue, we solve
\begin{equation}\small
\label{eq:design}
\begin{aligned}
\max_{V\in\mathbb R^{3L^2\times m}}\operatorname{tr}\big(V^\top M_{\mathcal E}V\big)
~~\text{s.t.}~V^\top\big(M_{\mathcal F}+\kappa I\big)V=I .
\end{aligned}
\end{equation}
The objective rewards encoder response, and the constraint measures every direction by its policy response.
The regularizer $\kappa>0$ keeps the problem well posed when some directions have almost no policy response.
The solution consists of the $m$ leading generalized eigenvectors $v_1,\ldots,v_m$,
\begin{equation}\small
\label{eq:eig}
\!\!\!\!M_{\mathcal E}v_j\!=\!\lambda_j\big(M_{\mathcal F}+\kappa I\big)v_j,~\lambda_1\!\ge\!\cdots\!\ge\!\lambda_m,~u_j\!\!=\!\!\sqrt{3b^2}\,\frac{\Phi v_j}{\|\Phi v_j\|},\!
\end{equation}
where $\lambda_j$ are the generalized eigenvalues and each $u_j$ has unit root-mean-square value per pixel (Appendix~\ref{sec:spectral_solution}).

\subsection{Adaptive Support}
\label{sec:support}

The third step places the noise on the tiles of each frame to which the reference policies respond least.
Following the bound in \cref{eq:quadform}, the response of frame $i$ at tile $p$ is
\begin{equation}\small
\label{eq:tileresp}
e_{i,p}=\sum_{\pi\in\mathcal F}\frac1{m\,\nu_\pi}\sum_{j=1}^{m}\big\|J_\pi(x_i)\,T_pu_j\big\|^2 ,
\end{equation}
where $\nu_\pi$ is the mean tile response of policy $\pi$ over a batch of frames, and one backward pass with a random output projection scores all tiles of a frame (Appendix~\ref{sec:support_bound}).

\vspace{1mm} \noindent \textbf{Support selection.}
For a coverage $\rho\in(0,1]$, the support keeps the $k=\lceil\rho P\rceil$ tiles with the smallest response,
\begin{equation}\small
\label{eq:mask}
s_{i,p}=\mathbb I\big[e_{i,p}\le e_{i,(k)}\big],
\end{equation}
where $e_{i,(k)}$ is the $k-$th smallest response in frame $i$, so the support minimizes $\sum_{p}s_{i,p}e_{i,p}$ among supports of $k$ tiles and $\rho=1$ recovers uniform tiling.

\vspace{1mm} \noindent \textbf{Support network.}
Since exact responses need a backward pass through every reference policy per frame, the generator computes them only on the frames $\mathcal I_0$ of 10\% of the episodes and predicts them elsewhere with a small convolutional network $h_\omega$.
Since \cref{eq:mask} depends only on the order of the responses within a frame, the network regresses their standardized logarithm,
\begin{equation}\small
\label{eq:learned_response}
\begin{aligned}
\min_{\omega}\sum_{i\in\mathcal I_0}\sum_{p=1}^{P}\big(h_\omega(x_i)_p-\frac{\log e_{i,p}-\mu_i}{\sigma_i}\big)^2,
\end{aligned}
\end{equation}
where $h_\omega(x_i)_p$ is the output of the network at tile $p$, and $\mu_i$ and $\sigma_i$ are the mean and standard deviation of $\log e_{i,p}$ over the tiles of frame $i$.
Every other frame needs one forward pass of $h_\omega$, and its support follows from \cref{eq:mask} with $h_\omega(x_i)$ in place of the exact responses.
The network also receives the camera index, and exact responses of sets protected earlier join the fit.
The parameters $\phi=(W^\star,c^\star,B,\Lambda,U,\omega)$ of the generator are computed in closed form except the network weights $\omega$.

\begin{algorithm}[t]
\caption{\method}
\label{alg:method}
\begin{algorithmic}[1]
\Require Demonstrations $\mathcal D$, reference policies $\mathcal F$, encoders $\mathcal E$, budget $\epsilon$, tile size $b$, band limit $L$, embedding size $m$, parameters $\kappa,\rho,\gamma$
\State Fit the regression and the embedding with \cref{eq:residual,eq:embedding}
\State Estimate the response matrices with \cref{eq:response,eq:response_ensemble}
\State Solve \cref{eq:design} for the patterns $U$ with \cref{eq:eig}
\State Compute exact tile responses on $\mathcal I_0$ with \cref{eq:tileresp} and fit $h_\omega$ with \cref{eq:learned_response}
\For{Each frame $i$}
  \State Predict tile responses with $h_\omega$ and select $s_i$ with \cref{eq:mask}
  \State Synthesize $\delta_i$ with \cref{eq:write}
\EndFor
\State Check both utility constraints on $\mathcal I_1$
\State If a constraint fails, adjust $\gamma,\rho,\kappa$ and repeat the affected steps
\Ensure The protected release $\widetilde{\mathcal D}$
\end{algorithmic}
\end{algorithm}

\subsection{Protected Release}
\label{sec:release}

The final step checks both utility constraints of \cref{eq:objective} on the saved images.
On a set $\mathcal I_1$ of 2,048 frames drawn at random from the release, we compute the error change and the drift of every reference policy with \cref{eq:offline}.
The release is accepted only if both stay within $\tau_E$ and $\tau_D$, and otherwise we adjust the gain $\gamma$, the coverage $\rho$ or the regularizer $\kappa$ and regenerate it.
The overall procedure is in \Cref{alg:method}.

%% file: sec/4_experiments.tex
\begin{table*}[t]
  \centering
  \small
  \caption{\textbf{Main results at $\epsilon=\frac{4}{255}$.} SR is the success of SmolVLA trained on each release (lower is stronger protection). $\Delta E$ and \textbf{Drift} are the error change and drift, \textit{i.e.,} $\frac{\big|E_\pi(\widetilde{\mathcal D})-E_\pi(\mathcal D)\big|}{E_\pi(\mathcal D)}$ and $\frac{\Delta_\pi}{E_\pi(\mathcal D)}$, with budgets of 2\% and 6\%. All values are reported as percentages.}
  \vspace{-3mm}
  \label{tab:main}
  \input{tables/main}
  \vspace{-1mm}
\end{table*}

\begin{figure*}[t]
  \centering

  \includegraphics[width=\textwidth]{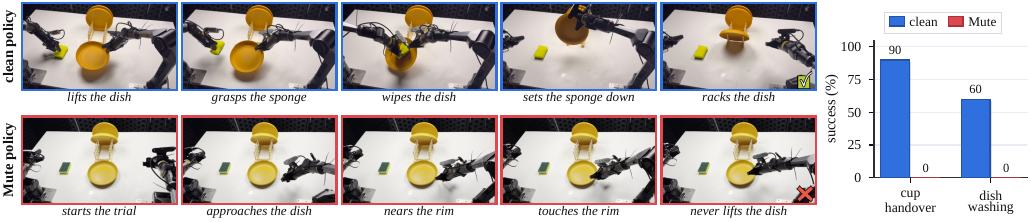}
  \vspace{-8mm}
  \caption{\textbf{Real-robot deployment.}
  Each row shows one dish-washing trial of $\pi_0$ trained on the
  clean demonstrations and on the \method release.
  Right, success of both policies on the robot.}
  \label{fig:robot}

  \vspace{1mm}

  \clipbox{0.27\textwidth{} 0pt 0.30\textwidth{} 0pt}{%
    \includegraphics[width=\textwidth]{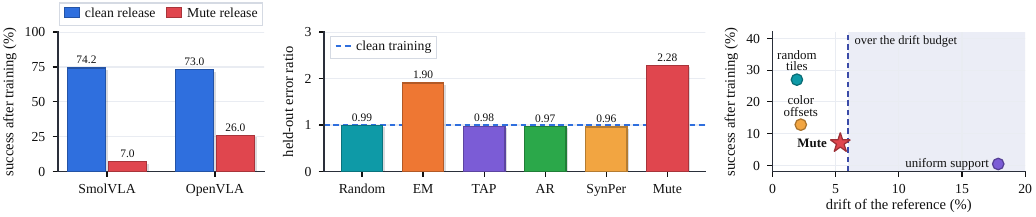}%
  }%
  \clipbox{0pt 0pt 0.73\textwidth{} 0pt}{%
    \includegraphics[width=\textwidth]{figs/results_transfer.pdf}%
  }%
  \clipbox{0.70\textwidth{} 0pt 0pt 0pt}{%
    \includegraphics[width=\textwidth]{figs/results_transfer.pdf}%
  }
\vspace{-8mm}
  \caption{\textbf{Real-world demonstrations, transfer and ablation.}
  Left: held-out error of SmolVLA trained on each SO-100 release
  relative to clean training, where a higher ratio means stronger protection.
  Middle: success of policies trained on LIBERO-Spatial releases.
  Right: reference drift against success when one component of
  \method changes.
  The corresponding numerical results are in
  \Cref{tab:real,tab:transfer,tab:ablation}.}
  \label{fig:transfer}

  \vspace{-4mm}
\end{figure*}

\section{Experiments}
\label{sec:experiments}


\noindent \textbf{Benchmarks.}
We use the Spatial, Object and Goal suites of LIBERO~\cite{liu2023libero}, each with 10 tasks and two $256\!\times\!256$ cameras.
Meta-World~\cite{yu2020metaworld} provides 10 MT10 tasks. RoboTwin~2.0~\cite{chen2025robotwin} provides 10 bimanual tasks with three cameras.
Each simulated policy is evaluated for 50 episodes per task in the clean simulator.
The real-world set merges 158 SO-100 episodes from SmolVLA~\cite{shukor2025smolvla}, with 10\% per task held out as clean test data.

\vspace{1mm}
\noindent \textbf{Policies.}
We fine-tune all parts of SmolVLA~\cite{shukor2025smolvla} for 20k steps with a batch size of 32.
To test transferability, we also train OpenVLA~\cite{kim2024openvla} on the same released pixels with its public fine-tuning recipe.
The real-robot experiment uses $\pi_0$ fine-tuned with LoRA.
To optimize the protected realease, the frozen reference polices are the public SmolVLA policy of each simulated benchmark, a SmolVLA policy trained on the clean SO-100 split and a $\pi_0$ policy trained on the clean robot demonstrations.

\vspace{1mm}
\noindent \textbf{Baselines.}
We compare \method with clean training, random sign noise, error-minimizing noise (EM)~\cite{huang2021unlearnable}, adversarial poisons (TAP)~\cite{fowl2021adversarial}, autoregressive noise (AR)~\cite{sandoval2022autoregressive} and synthetic perturbations (SynPer)~\cite{yu2022availability}.
Since these methods were designed for class labels, EM and TAP regress the embedding $z_i$ of \cref{eq:embedding} and AR and SynPer use 50 $k$-means clusters of $z_i$ as classes. 
More details are in Appendix~\ref{sec:details}.

\vspace{1mm}
\noindent \textbf{Metrics.}
Protection is the clean-environment success of the trained policy and on SO-100 its action error on held-out episodes.
Evaluation utility is the change of the action error and drift in percent of its clean error, \textit{i.e.,} $\Delta E\!=\!\frac{\big|\!E_\pi(\widetilde{\mathcal D})\!-\!E_\pi(\mathcal D)\!\big|}{E_\pi(\mathcal D)}$ \& $\textbf{Drift}\!\!=\!\!\frac{\Delta_\pi}{E_\pi\!(\mathcal D)}$, with budgets of 2\% and 6\%.

\vspace{1mm}
\noindent \textbf{Implementation details.}
We use $\epsilon=\frac{4}{255}$, tiles of $b=16$ pixels, $L=13$ frequencies per axis and $m=24$ embedding coordinates for a window of $H=50$ actions.
The patterns use $\kappa=0.3$, and the support keeps a share $\rho=0.5$ of the tiles at gain $\gamma=1$.
The network $h_\omega$ is fitted on exact responses of 10\% of the episodes.
Every configuration is trained once, and Appendix \ref{sec:details} lists the remaining details.

\subsection{Main Results}
\label{sec:main}

\begin{figure*}[t]
  \centering
  \includegraphics[width=\textwidth]{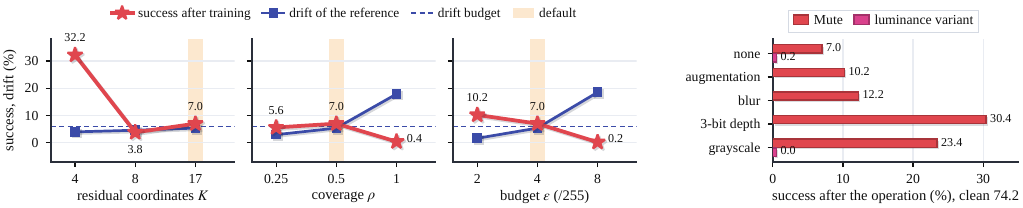}
  \vspace{-6mm}
  \caption{\textbf{Hyperparameters and robustness on LIBERO-Spatial.} Left: success after training and reference drift when one hyperparameter changes. Right: success after the data user applies an image operation to the release before training.  Numerical results: Tabs. \ref{tab:hyper}, \ref{tab:robust}.}
  \label{fig:analysis}
  \vspace{-2mm}
\end{figure*}

\noindent \textbf{Results across different benchmarks.}
\Cref{tab:main} tests whether a protected release lowers the success of trained policies while meeting both budgets.
On the three LIBERO suites, the \method release changes the reference error by at most 2.0\% and keeps the drift at or below 5.4\% of the error.
Training on it reduces success from 74.2\% to 7.0\% on Spatial, from 74.4\% to 0.0\% on Object and from 85.4\% to 0.4\% on Goal.
Among the unlearnable-example baselines within both budgets, AR is the strongest and lowers success to 35.6\% on Object and 27.5\% on Goal, but it leaves 72.1\% on Spatial.
EM and random noise leave between 46.2\% and 75.3\% success, and both exceed the two budgets on Spatial and Goal with error changes near 10\%.
TAP and SynPer leave success near clean training, and TAP exceeds the error budget on every suite.
On Meta-World \method reduces success from 62.6\% to 33.4\%, while the baselines stay between 46.9\% and 65.3\%.
On RoboTwin it perturbs all three cameras and reduces success from 78.4\% to 17.0\%, with an error change of \textminus{}0.1\% and a drift of 0.1\%.
No baseline lowers success below 57.3\% on RoboTwin.

\noindent \textbf{Results on real-world demonstrations.}
On the real-world SO-100 set, the policy trained on the \method release makes 2.28 times the held-out error of clean training, as shown in the left plot in \cref{fig:transfer}.
EM raises it by a factor of 1.90, and the other perturbations leave it within 4\% of clean training.

\vspace{1mm}
\noindent \textbf{Results on real-robot deployment.}
We deploy \method on a bimanual robot with three cameras and 14-dimensional joint actions, using 100 demonstrations for each of three tasks.
The protected release is built with the SigLIP encoder of $\pi_0$ and luminance patterns on a quarter of the tiles, and it changes the error of a frozen $\pi_0$ reference by 0.44\% with a drift of 0.56\%.
We separately train $\pi_0$ on the clean demonstrations and the protected release, and run both policies on the robot.
The clean policy succeeds in 90\% of the trials on cup handover and 60\% on dish washing, while the \method policy succeeds in no trial on either task, as shown in \cref{fig:robot}.
In the trial shown, the \method policy never lifts the dish or grasps the sponge, so the wiping never starts.

\vspace{1mm}\noindent\textbf{Visualizations of perturbations.} Appendix \cref{fig:supp_vis} compares clean and protected frames, showing that the perturbations introduce little visible change.



\begin{table}[t]
  \centering
  \small
  \caption{\textbf{Varying surrogate models on LIBERO-Spatial.}
  The victim is SmolVLA. The self-trained reference fine-tunes public SmolVLA on the target clean demonstrations.
  Utility is measured on the public reference. All values are reported as percentages.}
  \vspace{-3mm}
  \label{tab:surrogate}
  \input{tables/surrogate}
  \vspace{-4mm}
\end{table}

\subsection{Transferability}
\label{sec:transfer}
Data owners cannot anticipate downstream policy architectures. We therefore vary both the victim policy and the surrogate models used to construct the release.

\vspace{1mm}\noindent\textbf{Varying victim policies.}
The middle plot of \cref{fig:transfer} compares policies trained on the same LIBERO-Spatial release optimized with SmolVLA.
OpenVLA achieves 26.0\% success on the \method release versus 73.0\% on the clean release, demonstrating transfer to a strong different policy.

\vspace{1mm}\noindent\textbf{Varying surrogate models.}
With SmolVLA as the victim, we vary the construction reference or encoder, as shown in \cref{tab:surrogate}.
Both surrogate substitutions retain effective protection, demonstrating strong transfer of \method across different surrogate models.

\subsection{Ablation Study}
\label{sec:ablation}

The right plot of \cref{fig:transfer} changes one component at a time, and only the full method reaches low success inside the drift budget. Detailed numerical results are in \Cref{tab:ablation} of the Appendix.
Tiling the same patterns over the whole image lowers success to 0.4\%, but it raises the reference error by 15.1\% and the drift to 17.9\%.
The null-response patterns are needed for protection, since random tile patterns leave 27.0\% success and per-patch color offsets leave 12.8\%.

\subsection{Hyperparameters and Robustness}
\label{sec:hyperparameters}

\noindent \textbf{Hyperparameters.}
With four residual coordinates the release leaves 32.2\% success, and from eight coordinates on success stays below 8\%, as shown in \cref{fig:analysis}.
A coverage of 0.25 keeps 5.6\% success with a drift of 3.0\%, while full coverage reaches 0.4\% with a drift of 17.9\%.
The pixel budget has the largest effect, with 10.2\% success and 1.6\% drift at $\epsilon=\frac{2}{255}$ and 0.2\% success but 18.5\% drift at $\epsilon=\frac{8}{255}$.

\vspace{1mm}
\noindent \textbf{Robustness.}
\label{sec:defenses}
The right plot of \cref{fig:analysis} applies image operations before training.
Augmentation and Gaussian blur leave the protection largely intact with 10.2\% and 12.2\% success, while 3-bit depth recovers 30.4\% and grayscale conversion 23.4\% because it removes the color component of the patterns.
A variant restricted to luminance and to the lowest frequencies of $32\times32$ tiles meets both budgets and reduces success to 0.2\% without any operation and to 0.0\% after grayscale conversion.

\subsection{Further Analysis}
\label{sec:analysis}

\noindent \textbf{Training dynamics.}
We track the policy's action prediction error on both protected and clean frames during training, as shown in  Appendix \cref{fig:support} and \Cref{tab:dynamics}.
After 10k steps, the policy trained on the \method release achieves an error of 0.031 on protected frames but 0.785 on clean frames.
This gap indicates that fitting the protected demonstrations does not translate into accurate prediction without the perturbations, consistent with early reliance on the injected shortcut.




\vspace{1mm}
\noindent \textbf{Computational efficiency of the generator and comparisons with baselines.}
Exact tile responses require backward passes through the reference policies for every frame.
The support network learns from exact responses on 10\% of episodes and predicts scores elsewhere, reducing support scoring from 137 to 42 minutes, including exact-response computation and fitting (Appendix \cref{tab:support}).
This provides a $3.3\times$ speedup while the resulting release meets both evaluation budgets.
Fitting on only 1\% of episodes still yields 1.4\% victim success (Appendix \cref{tab:share}).
The full pipeline takes approximately 69 minutes on LIBERO-Spatial, a practical one-time preprocessing cost comparable to that of optimization-based baselines (Appendix~\ref{sec:cost}).

%% file: tables/main.tex
\renewcommand{\arraystretch}{1.08}
\setlength{\tabcolsep}{6pt}
\begin{tabular}{l||ccc|ccc|ccc|ccc|ccc}
\toprule
\rowcolor{tabhead} & \multicolumn{3}{c|}{\textbf{LIBERO-Spatial}} & \multicolumn{3}{c|}{\textbf{LIBERO-Object}} & \multicolumn{3}{c|}{\textbf{LIBERO-Goal}} & \multicolumn{3}{c|}{\textbf{Meta-World}} & \multicolumn{3}{c}{\textbf{RoboTwin}} \\
\rowcolor{tabhead} \multirow{-2}{*}{\textbf{Methods}} & SR & $\Delta E$ & \textbf{Drift} & SR & $\Delta E$ & \textbf{Drift} & SR & $\Delta E$ & \textbf{Drift} & SR & $\Delta E$ & \textbf{Drift} & SR & $\Delta E$ & \textbf{Drift} \\
\midrule
Clean & 74.2 & -- & -- & 74.4 & -- & -- & 85.4 & -- & -- & 62.6 & -- & -- & 78.4 & -- & -- \\
\arrayrulecolor{black!45}
\midrule
\arrayrulecolor{black}
\rowcolor{tabzebra} Random noise & 56.3 & 10.4 & 12.9 & 54.9 & 1.2 & 2.9 & 75.3 & 10.5 & 12.3 & 63.6 & 0.3 & 1.8 & 72.5 & 2.1 & 0.3 \\
EM~\cite{huang2021unlearnable} & 65.9 & 10.7 & 14.0 & 56.4 & 1.2 & 2.8 & 46.2 & 9.6 & 11.8 & 59.6 & 0.4 & 3.0 & 63.7 & 3.5 & 0.7 \\
\rowcolor{tabzebra} TAP~\cite{fowl2021adversarial} & 70.6 & 23.6 & 25.9 & 55.9 & 3.4 & 5.3 & 75.7 & 11.1 & 13.4 & 65.3 & 1.6 & 5.8 & 71.5 & 9.1 & 3.1 \\
AR~\cite{sandoval2022autoregressive} & 66.1 & 0.4 & 1.6 & 35.6 & 0.0 & 0.5 & 27.5 & 0.2 & 1.4 & 59.4 & 0.0 & 0.2 & 57.3 & 0.1 & 0.0 \\
\rowcolor{tabzebra} SynPer~\cite{yu2022availability} & 65.3 & 1.5 & 2.6 & 61.5 & 0.1 & 0.6 & 65.9 & 1.0 & 2.2 & 46.9 & 0.4 & 0.9 & 65.0 & 0.2 & 0.0 \\
\midrule
\rowcolor{tabours} \textbf{\method (ours)} & \textbf{7.0} & \textbf{2.0} & \textbf{5.4} & \textbf{0.0} & \textbf{0.2} & \textbf{1.1} & \textbf{0.4} & \textbf{1.0} & \textbf{3.0} & \textbf{33.4} & \textbf{0.3} & \textbf{1.3} & \textbf{17.0} & \textbf{\textminus{}0.1} & \textbf{0.1} \\
\bottomrule
\end{tabular}

%% file: tables/surrogate.tex
\setlength{\tabcolsep}{4pt}
\renewcommand{\arraystretch}{1.08}
\begin{tabular}{lccc}
\toprule
\rowcolor{tabhead} \textbf{Models used to build the release} & \textbf{SR} & \textbf{$\Delta E$} & \textbf{Drift} \\
\midrule
Self-trained SmolVLA reference & 2.8 & 3.8 & 6.4 \\
\rowcolor{tabzebra} SigLIP encoder & 23.0 & 1.5 & 4.1 \\
\rowcolor{tabours} \textbf{Public reference and SmolVLA encoder} & \textbf{7.0} & \textbf{2.0} & \textbf{5.4} \\
\bottomrule
\end{tabular}

%% file: sec/5_conclusion.tex
\section{Conclusion}
\label{sec:conclusion}
\vspace{-1mm}


Sharing robot demonstrations as evaluation benchmarks supports reproducible policy comparisons, but also exposes costly data to unauthorized training.
We addressed this challenge with \method, guided by empirical observations and theoretical analysis.
\method encodes continuous action windows into training shortcuts through image perturbations, then selects noise patterns and spatial support to limit their effects on frozen-policy predictions.
Experiments across simulated benchmarks, real-world demonstrations and real-robot deployment show strong protection with preserved evaluation fidelity.
These results show the potential to retain the benchmarking value of robot demonstrations while reducing their utility for training.


\vspace{1mm}
\noindent\textbf{Societal impact.}
This approach gives data owners control over training use, but developers who unknowingly train on protected releases may deploy unreliable policies that endanger people or property.
The perturbations could also be misused to degrade datasets the publisher does not own.
Owners should label protected releases, dataset hosts should track provenance, and developers should validate policies on clean demonstrations before deployment.

%% file: sec/6_llm_usage.tex
\section*{LLM Usage Statement}
We used LLMs to assist with writing and code implementation.
For the code, it implemented the part of experiment scripts and the scripts that generate the tables and figures.
The authors directed the work and take full responsibility for the usage.

%% file: sec/X_suppl.tex


\section{Training Objectives}
\label{sec:training_losses}

The sample loss $\mathcal L$ in \cref{eq:bc} depends on the action head.
Protection changes only the image in $\tilde o_i$ and keeps the action targets and the padding of the windows.

\paragraph{Flow matching.}
SmolVLA~\cite{shukor2025smolvla} and $\pi_0$~\cite{black2024pi0} learn a velocity field $v_\theta$ by flow matching~\cite{lipman2023flow}.
Following the noise-to-action convention of their LeRobot implementations, the noisy window at time $t\in[0,1]$ is
\begin{equation}
\label{eq:flow_path}
A_i^{t}=(1-t)\,A_i+t\,\eta,\qquad \eta\sim\mathcal N(0,I),
\end{equation}
where $\eta$ is Gaussian noise of the same shape as $A_i$.
The sample loss is
\begin{equation}
\label{eq:flow_loss}
\mathcal L(\theta;\tilde o_i,A_i)
=\mathbb E_{t\sim p_t,\,\eta}\big\|v_\theta(A_i^{t},t,\tilde o_i)-(\eta-A_i)\big\|^2 ,
\end{equation}
where $p_t$ is the time distribution of the training recipe.
At inference the policy starts from noise at $t=1$ and integrates the velocity field toward $t=0$.
The velocity regression therefore differs from the action error in \cref{eq:offline}, which evaluates the integrated actions.

\paragraph{Action tokens.}
OpenVLA~\cite{kim2024openvla} discretizes every action dimension into bins and predicts the resulting tokens $c_{i,1},\ldots,c_{i,N_i}$ one after another.
Its sample loss is the negative log-likelihood
\begin{equation}
\label{eq:token_loss}
\mathcal L(\theta;\tilde o_i,A_i)
=-\frac1{N_i}\sum_{j=1}^{N_i}\log p_\theta\big(c_{i,j}\mid c_{i,1},\ldots,c_{i,j-1},\tilde o_i\big),
\end{equation}
where $p_\theta$ is the token distribution of the language model.
The tokenizer and the prediction horizon follow the native recipe.
Decoded tokens are mapped back to standardized actions for evaluation.

\section{Analysis and Derivations}
\label{sec:proofs}

\subsection{Local Prediction Response}
\label{sec:proof_inert}

Fix the sampling noise and every input other than the image.
Taylor's theorem gives $\pi(\tilde o)-\pi(o)=J_\pi\delta+\varrho$, where the remainder satisfies $\|\varrho\|\le c_0\|\delta\|^2$ for a constant $c_0$ near $x$.
Hence
\begin{equation}
\begin{aligned}
\|\pi(\tilde o)-\pi(o)\|^2
&=\|J_\pi\delta\|^2+2\langle J_\pi\delta,\varrho\rangle+\|\varrho\|^2\\
&=\|J_\pi\delta\|^2+O\big(\|\delta\|^3\big).
\end{aligned}
\end{equation}
Substituting $\delta=\sum_{p=1}^{P}s_p\,T_pu$ gives $J_\pi\delta=\sum_{p}s_pJ_\pi T_pu$, which proves the equality in \cref{eq:quadform}.
This sum has $k$ nonzero terms, and the Cauchy-Schwarz inequality gives
\begin{equation}
\Big\|\sum_{p=1}^{P}s_p\,J_\pi T_pu\Big\|^2\le k\sum_{p=1}^{P}s_p\,\|J_\pi T_pu\|^2 ,
\end{equation}
which proves the bound.
The interactions between selected tiles are dropped only in the bound.

\subsection{Least-Squares Weighting}
\label{sec:proof_weight}

Consider a centered scalar action $y$ with unit variance.
Let $f_1=y+\xi_1$ be a feature of the original observation and $f_2=y+\xi_2$ a feature carried by the perturbation.
The noise terms $\xi_1$ and $\xi_2$ have zero mean and variances $\sigma_1^2>0$ and $\sigma_2^2\ge0$, and $y$, $\xi_1$ and $\xi_2$ are independent.

\begin{proposition}[Weight of a predictive perturbation]
\label{prop:weight}
The minimizer of $\mathbb E[(\beta_1f_1+\beta_2f_2-y)^2]$ is
\begin{equation}
\label{eq:weights}
\begin{aligned}
\beta_1&=\frac{\sigma_2^2}{\sigma_1^2+\sigma_2^2+\sigma_1^2\sigma_2^2},\\
\beta_2&=\frac{\sigma_1^2}{\sigma_1^2+\sigma_2^2+\sigma_1^2\sigma_2^2}.
\end{aligned}
\end{equation}
If $\sigma_2^2<\sigma_1^2$, then $\beta_2>\beta_1$.
On clean inputs the perturbation feature is absent, and the error becomes
\begin{equation}
\label{eq:clean_error}
\mathbb E\big[(\beta_1f_1-y)^2\big]=(1-\beta_1)^2+\beta_1^2\sigma_1^2 .
\end{equation}
This error increases as $\sigma_2^2$ decreases and equals $1$ at $\sigma_2^2=0$.
\end{proposition}

\begin{proof}
Independence gives the feature covariance and the cross-covariance with the target,
\begin{equation}
\Sigma_f=
\begin{pmatrix}
1+\sigma_1^2&1\\
1&1+\sigma_2^2
\end{pmatrix},
\qquad
\mathbb E\!\left[\begin{pmatrix}f_1\\f_2\end{pmatrix}y\right]
=\begin{pmatrix}1\\1\end{pmatrix}.
\end{equation}
The determinant of $\Sigma_f$ is positive, so the normal equations have the unique solution $(\beta_1,\beta_2)^\top=\Sigma_f^{-1}(1,1)^\top$.
Inverting $\Sigma_f$ gives \cref{eq:weights}.
Writing the clean error in \cref{eq:clean_error} as $\psi(\beta_1)$, we obtain
\begin{equation}
\begin{aligned}
\psi'(\beta)&=2\big((1+\sigma_1^2)\beta-1\big),\qquad
0\le\beta_1<\frac1{1+\sigma_1^2},\\
\frac{\partial\beta_1}{\partial\sigma_2^2}
&=\frac{\sigma_1^2}{\big(\sigma_1^2+(1+\sigma_1^2)\sigma_2^2\big)^2}>0 .
\end{aligned}
\end{equation}
The error therefore decreases in $\beta_1$, and $\beta_1$ increases with $\sigma_2^2$.
At $\sigma_2^2=0$ we have $\beta_1=0$ and $\psi(0)=1$.
\end{proof}

This model isolates the reliability of a feature under linear fitting.
It does not describe the training dynamics of a nonlinear VLA or equate action error with task failure.
A larger embedding carries more action information, but whether it strengthens protection must be tested empirically.

\subsection{Pattern Optimization}
\label{sec:spectral_solution}

Let $Q=M_{\mathcal F}+\kappa I$, which is positive definite for $\kappa>0$, and assume $m\le3L^2$.
For a single pattern, \cref{eq:design} maximizes the generalized Rayleigh quotient
\begin{equation}
\label{eq:rayleigh}
\max_{v\ne0}\ \frac{v^\top M_{\mathcal E}v}{v^\top Qv}.
\end{equation}
For several patterns, the constraint $V^\top QV=I$ prevents the columns from choosing the same direction.
With $\widetilde M=Q^{-1/2}M_{\mathcal E}Q^{-1/2}$ and $\widetilde V=Q^{1/2}V$, the problem becomes
\begin{equation}
\max_{\widetilde V^\top\widetilde V=I}\ \operatorname{tr}\big(\widetilde V^\top\widetilde M\widetilde V\big).
\end{equation}
The optimum is the sum of the $m$ largest eigenvalues of the symmetric matrix $\widetilde M$ and is attained by their eigenvectors.
Mapping back with $V=Q^{-1/2}\widetilde V$ gives the generalized eigenvectors in \cref{eq:eig}.
The scaling of $u_j$ does not preserve the constraint on $V$.
The optimization therefore fixes the pattern directions, while the gain $\gamma$ and the clipping in \cref{eq:write} determine the released amplitude.

\paragraph{Response estimator.}
For a model $f$, a frame $i$ and a standard Gaussian vector $\zeta$ of the output dimension, one backward pass gives
\begin{equation}
\begin{aligned}
g_{f,i}(\zeta)&=T^\top J_f(x_i)^\top\zeta,\\
\mathbb E_\zeta\big[g_{f,i}(\zeta)\,g_{f,i}(\zeta)^\top\big]&=T^\top J_f(x_i)^\top J_f(x_i)\,T .
\end{aligned}
\end{equation}
We average $g_{f,i}(\zeta)g_{f,i}(\zeta)^\top$ over 1,024 random frames and four draws of $\zeta$ per frame, and a policy that samples its actions draws fresh sampling noise for each projection.
The estimate is unbiased for $G_f$ and positive semidefinite.
The trace normalization and the eigenvector selection are nonlinear, so the estimated patterns are not unbiased in general.

\subsection{Tile Response and Support}
\label{sec:support_bound}

\paragraph{Response of the synthesized noise.}
Let $\delta_i^{\circ}=\alpha\sum_{p}s_{i,p}T_pUz_i$ with $\alpha=\gamma\epsilon/\sqrt m$ be the perturbation of \cref{eq:write} before clipping, and let $k_i=\sum_p s_{i,p}$.
For a policy $\pi$, define the single-policy tile response $e_{\pi,i,p}=m^{-1}\sum_{j}\|J_\pi(x_i)T_pu_j\|^2=m^{-1}\|J_\pi(x_i)T_pU\|_F^2$, where $\|\cdot\|_F$ is the Frobenius norm.
The Cauchy-Schwarz inequality and the Frobenius bound give
\begin{equation}
\label{eq:support_bound}
\begin{aligned}
\|J_\pi(x_i)\,\delta_i^{\circ}\|^2
&\le\alpha^2k_i\sum_{p=1}^{P}s_{i,p}\,\|J_\pi(x_i)T_pUz_i\|^2\\
&\le\gamma^2\epsilon^2k_i\,\|z_i\|^2\sum_{p=1}^{P}s_{i,p}\,e_{\pi,i,p} .
\end{aligned}
\end{equation}
For a fixed coverage, embedding and patterns, choosing tiles with a small response reduces this bound.
With several cameras the same bound holds with $k_i$ counting the selected tiles of all views.
Clipping and rounding can change the direction of the perturbation, so the bound is not a guarantee on the drift of the saved images.

\paragraph{Tile-response estimator.}
The image gradient $J_\pi(x_i)^\top\zeta$ of one random projection scores every tile and pattern,
\begin{equation}
\label{eq:tile_estimator}
\widehat e_{\pi,i,p}=\frac1{mR}\sum_{l=1}^{R}\sum_{j=1}^{m}\big\langle J_\pi(x_i)^\top\zeta_l,\,T_pu_j\big\rangle^2 ,
\end{equation}
where $\zeta_1,\ldots,\zeta_R$ are independent standard Gaussian vectors.
The estimate is unbiased for $e_{\pi,i,p}$.
The normalizer $\nu_\pi$ in \cref{eq:tileresp} is the mean of $\widehat e_{\pi,i,p}$ over the tiles, cameras and frames of the current batch.

\paragraph{Support selection.}
Suppose a support of $k$ tiles contains a tile with a larger response than an unselected tile.
Exchanging the two tiles keeps the size and reduces $\sum_p s_{i,p}e_{i,p}$.
Repeated exchanges lead to the $k$ smallest responses, which is the rule in \cref{eq:mask}.
Tiles tied with the $k$th response are all kept, so the realized size can exceed $k$.

\subsection{Prediction Drift and Evaluation Error}
\label{sec:utility_bound}

The two utility constraints in \cref{eq:objective} control different quantities.
Expanding the squared error and applying the Cauchy-Schwarz inequality over frames gives
\begin{equation}
\label{eq:utility_bound}
\begin{aligned}
E_\pi(\widetilde{\mathcal D})-E_\pi(\mathcal D)
&=\frac1{nHd}\sum_{i=1}^{n}\Big(2\big\langle\pi(o_i)-A_i,\,\pi(\tilde o_i)-\pi(o_i)\big\rangle\\
&\qquad\qquad+\big\|\pi(\tilde o_i)-\pi(o_i)\big\|^2\Big),\\
\big|E_\pi(\widetilde{\mathcal D})-E_\pi(\mathcal D)\big|
&\le2\sqrt{E_\pi(\mathcal D)\,\Delta_\pi}+\Delta_\pi ,
\end{aligned}
\end{equation}
where $\langle\cdot,\cdot\rangle$ is the inner product of flattened matrices.
A drift bound alone therefore does not imply the relative error tolerance $\tau_E$.
Conversely, a small error change can hide large drift because the cross term can be negative.
We therefore check both constraints directly.

\section{Additional Implementation Details}
\label{sec:details}

\paragraph{Trajectory statistics.}
The regression in \cref{eq:residual} and the residual covariance use all frames of the protected set.
Each whitening standard deviation is floored at $10^{-6}$.
Padded steps at the end of an episode repeat the last action when the embedding is fitted, while the offline metrics exclude them.
Every benchmark uses windows of $H=50$ actions, the chunk size of SmolVLA and $\pi_0$.
The embedding has $m=24$ coordinates, so $K=24-d$ is 17 on LIBERO and 20 on Meta-World while it is 10 on RoboTwin and 18 on SO-100.

\paragraph{Frame sets.}
The response matrices of \cref{eq:response} use 1,024 frames drawn at random from the protected set.
The exact tile responses use every frame of a random 10\% of its episodes, which form $\mathcal I_0$.
The utility check uses 2,048 frames $\mathcal I_1$ drawn at random and independently of both sets.
On the LIBERO suites, 9.5\% to 11.2\% of the frames of $\mathcal I_1$ lie in episodes of $\mathcal I_0$, and 1.3\% to 1.9\% also belong to the frames of the response matrices.

\paragraph{Differentiable reference outputs.}
The construction uses a continuous-action SmolVLA reference.
Its Jacobian differentiates through the sampled action computation with the sampling noise fixed.
A reference with discrete action tokens would require a differentiable output surrogate, such as the soft token expectation used for VLA attacks~\cite{wang2025vlaadversarial}.
Held-out models used only for utility evaluation do not require such a surrogate.

\paragraph{Gain and clipping.}
All releases in the main results use $\gamma=1$, which sets the root-mean-square value of the tile signal in \cref{eq:write} to about $\epsilon$.
On LIBERO-Spatial the support covers 49.9\% of the pixels of each camera.
Of the perturbed values, 33\% to 35\% reach the budget $\epsilon$, and their mean magnitude is $2.4/255$.
Meta-World uses tiles of $b=30$ pixels on its $480\times480$ frames.
\paragraph{Model inputs.}
The patterns are built on the $256\times256$ LIBERO frames that every policy receives.
SmolVLA resizes them to $512\times512$, so a tile of $b=16$ pixels covers $2\times2$ patches of its vision encoder.
$\pi_0$ and OpenVLA resize them to $224\times224$, where a tile covers one patch of 14 pixels.
The encoder response of SmolVLA uses the visual tokens of its image embedding, and the SigLIP surrogate uses the patch tokens of SigLIP-So400m/14 at $224\times224$.
\paragraph{Baselines.}
The four unlearnable-example baselines were proposed for image classification, where every training image carries one of a few class labels.
In a demonstration dataset every frame is instead paired with a continuous chunk of future actions, so none of the four methods applies directly.
We therefore adapt each method to this setting and keep its original noise construction wherever the class label does not enter.
In all four adaptations the label of frame $i$ is the trajectory embedding $z_i\in\mathbb{R}^m$ of \cref{eq:embedding}, which summarizes its action chunk.
Every camera image of a protected frame receives its own perturbation, which is scaled to $\|\delta_i\|_\infty=\epsilon$ as in \method.

\noindent\textit{EM.}
The original method trains a classifier on the perturbed images and minimizes its cross-entropy loss with respect to sample-wise noise.
We replace the cross-entropy loss with the squared error of a surrogate regressor $f_\theta$ that predicts the embedding,
\begin{equation}
\min_{\theta}\ \frac{1}{n}\sum_{i=1}^{n}\ \min_{\|\delta_i\|_\infty\le\epsilon}\ \big\|f_\theta(x_i+\delta_i)-z_i\big\|_2^2 ,
\end{equation}
where $x_i$ holds the clean camera images of frame $i$ and $n$ is the number of frames.
As in the original method, the optimization alternates updates of $\theta$ with signed gradient steps on every $\delta_i$.
It stops once the training error falls below 1\% of the total variance of the embedding, which takes the place of the original criterion of 99\% training accuracy.

\noindent\textit{TAP.}
The original method trains a classifier on clean images and turns each image into an adversarial example of another class, which a fixed shift of the labels selects.
We instead train the regressor $f_\theta$ on clean frames and push its prediction toward the embedding of another frame,
\begin{equation}
\delta_i=\operatorname*{arg\,min}_{\|\delta\|_\infty\le\epsilon}\ \big\|f_\theta(x_i+\delta)-z_{\pi(i)}\big\|_2^2 ,
\end{equation}
where $\pi$ is a fixed random cyclic permutation of the frames that takes the place of the label shift.
The minimization uses signed gradient steps with random restarts as in the original method.

\noindent\textit{AR and SynPer.}
Both methods assign one fixed noise pattern to each class and add the pattern of its class to every image.
Continuous labels have no classes, so we draw one pattern $n_j$ per embedding coordinate and weight it by that coordinate,
\begin{equation}
\delta_i=\epsilon\,\frac{s_i}{\|s_i\|_\infty},\qquad s_i=\sum_{j=1}^{m} z_{i,j}\,n_j ,
\end{equation}
where $z_{i,j}$ is the $j$-th coordinate of $z_i$.
When $z_i$ is a one-hot class indicator, $s_i$ reduces to the class pattern of the original methods.
Frames with similar action chunks thus receive similar perturbations, which preserves the link between label and noise that both methods rely on.
AR takes each $n_j$ as one sample of its own autoregressive process, whose coefficients we select at our image resolution with the search procedure of the original work.
SynPer takes each $n_j$ as a random vertex of a hypercube and adds to $s_i$ the Gaussian scatter of its synthetic data generator.
Each entry of the resulting pattern then covers a block of $8\times8$ pixels.

\paragraph{Training recipes.}
SmolVLA follows the LeRobot LIBERO recipe with full fine-tuning for 20k steps at batch size 32.
OpenVLA is fine-tuned with LoRA of rank 32 at a learning rate of $5\cdot10^{-4}$ for 10k steps at batch size 16 with its official image augmentation.

\paragraph{Real-robot deployment.}
The robot release follows the pipeline of the main paper with the following settings.
The reference is a $\pi_0$ policy whose action expert is trained for 3k steps on the clean demonstrations, and the encoder is the vision tower of $\pi_0$.
The luminance patterns use tiles of 28 pixels on the $224\times224$ input and response matrices from 512 frames.
Exact responses are computed on every second frame of 5\% of the episodes.
The release uses $\epsilon=4/255$, $\rho=0.25$ and $\gamma=0.5$ on all three cameras.
The protected policy is $\pi_0$ with LoRA, trained for 20k steps at a batch size of 32.

\section{Visualizations}
\label{sec:visual}

\begin{figure*}[t]
  \centering
  \includegraphics[width=\textwidth]{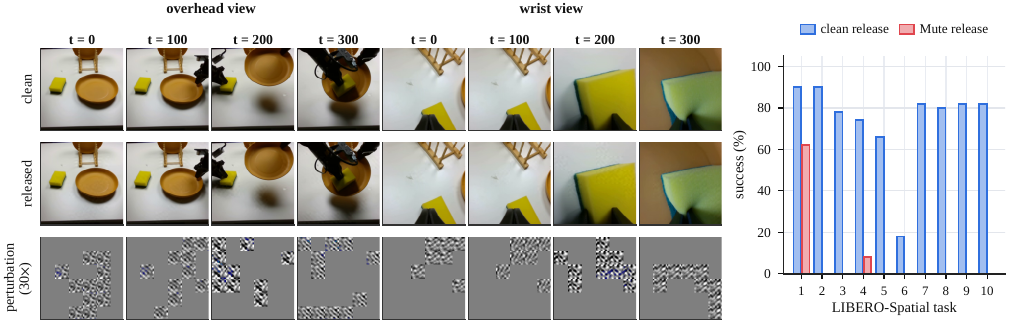}
  \caption{\textbf{Released frames and per-task success.} Left: one real dish-washing demonstration of the $\pi_0$ release on two cameras as clean frame, released frame and perturbation amplified 30 times around mid-gray. Right: success of SmolVLA trained on the clean and on the \method release for each of the ten LIBERO-Spatial tasks over 50 episodes each.}
  \label{fig:supp_vis}
\end{figure*}

\begin{figure}[t]
  \centering
\includegraphics[width=\columnwidth]{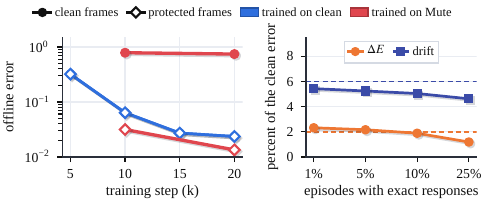}
  \caption{\textbf{Training dynamics and exact responses on LIBERO-Spatial.} Left, offline error of SmolVLA checkpoints on clean and protected frames, where the two curves of the policy trained on clean data coincide. Right, error change and drift of the release when the support network is fitted on exact tile responses of a share of the episodes, with the budgets of 2\% and 6\% dashed.}
  \label{fig:support}
\end{figure}


\Cref{fig:supp_vis} presents release visualizations and per-task protection results.
The left panel compares clean and protected frames from a real dish-washing demonstration captured by overhead and wrist cameras.
At $\epsilon=4/255$, the protected frames remain visually similar to the originals; the perturbations are amplified $30\times$ to reveal their spatial support and variation across frames.
The right panel reports per-task success on LIBERO-Spatial.
Eight of the ten tasks have zero success after training on the protected release, while the first task retains 62\% success and accounts for most of the 7.0\% average.

\section{Additional Results}
\label{sec:more_results}

The tables of this section list the numbers behind \cref{fig:transfer} of the Appendix, \cref{fig:analysis,fig:support} of the main paper together with the releases built with other models and the support of the generator.

\begin{table}[t]
  \centering
  \scriptsize
  \caption{\textbf{Real-world SO-100 demonstrations.} Held-out error is the action error of SmolVLA trained on each release, measured on clean held-out episodes, and Ratio divides it by the error of clean training. $\Delta E$ and Drift are measured on the frozen reference in percent of its clean error.}
  \label{tab:real}
  \input{tables/real}
\end{table}

\begin{table}[t]
  \centering
  \scriptsize
  \caption{\textbf{Policies trained on the same LIBERO-Spatial release.} Clean-environment success in percent.}
  \label{tab:transfer}
  \input{tables/transfer}
\end{table}

\begin{table}[t]
  \centering
  \scriptsize
  \caption{\textbf{Ablation on LIBERO-Spatial.} Each row changes one component of \method. SR is the success of SmolVLA trained on the release, and $\Delta E$ and Drift are measured on the frozen reference in percent of its clean error.}
  \label{tab:ablation}
  \input{tables/ablation}
\end{table}



\begin{table}[t]
  \centering
  \scriptsize
  \caption{\textbf{Robustness to image operations on LIBERO-Spatial.} Success of SmolVLA trained on the release after the data user applies each operation. The luminance variant restricts the patterns to luminance and DCT indices below 8 on $32\times32$ tiles.}
  \label{tab:robust}
  \input{tables/robust}
\end{table}

\begin{table}[t]
  \centering
  \scriptsize
  \caption{\textbf{Hyperparameters on LIBERO-Spatial.} One parameter changes at a time, and the default setting is shaded.}
  \label{tab:hyper}
  \input{tables/hyper}
\end{table}

\begin{table}[t]
  \centering
  \scriptsize
  \caption{\textbf{Share of episodes with exact tile responses.} The support network of the generator is fitted on exact responses of the given share of LIBERO-Spatial episodes. Overlap and admitted response are measured on held-out frames as in \cref{tab:support}. All releases are evaluated in one run, in which the default release (10\%) has an error change of 1.9\% and a drift of 5.0\% against 2.0\% and 5.4\% in the main results from an evaluation on a different GPU type.}
  \label{tab:share}
  \input{tables/share}
\end{table}

\begin{table}[t]
  \centering
  \scriptsize
  \caption{\textbf{Support of the generator on LIBERO-Spatial.} Columns give the support source, a camera-wise mean map, a network fitted on other suites only, the default network of the generator fitted on 10\% of the Spatial episodes as well, and exact estimates on every frame. Rows give the overlap with the exact support and the exact response admitted per tile (uniform tiling is 1) on held-out frames, the reference-policy error change and drift of the release built with each support, the success of SmolVLA trained on it, and the time to score the suite on one GPU. Dashes mark releases we did not build or train on.}
  \label{tab:support}
  \input{tables/support}
\end{table}

\paragraph{Learned tile response.}
\Cref{tab:support} compares the default support of the generator with exact estimates on every frame.
The default network $h_\omega$ is fitted on exact estimates of 10\% of the Spatial episodes together with those of the Object and Goal sets.
On the held-out frames the prediction places 81.5\% of its tiles in the exact support.
It admits 0.16 of the mean tile response, against 0.07 for the exact support and 0.37 for a camera-wise mean map.
The release built from the predicted support changes the reference error by 2.0\% with a drift of 5.4\% of the error, and the policy trained on it reaches 7.0\% success against 8.8\% with exact estimates.
Both supports meet the drift budget of 6\%, and the predicted support uses more of it.
The generator scores the suite in 42 minutes including the exact estimates and the fitting, against 137 minutes for exact estimates on every frame, and the saving grows with the size of the release.
A network fitted without any Spatial estimate still covers 75.2\% of the exact support, but its drift of 6.3\% exceeds the budget.
A small set of exact estimates on the protected set therefore remains part of the default.

\begin{table}[t]
  \centering
  \scriptsize
  \caption{\textbf{Offline error during training on LIBERO-Spatial.} Each SmolVLA checkpoint is evaluated as a frozen policy on clean and \method-protected frames. Drift is the mean squared change of its prediction between the two frames.}
  \label{tab:dynamics}
  \input{tables/dynamics}
\end{table}

\section{Computational Cost}
\label{sec:cost}

\Cref{tab:cost} summarizes runtimes for
protecting the 52,970 frames of LIBERO-Spatial on an A6000.
The reported cost of \method is 68.9 minutes, or 78.0 milliseconds
per frame, including pattern construction, support scoring and
release writing.
These figures cover the overall release-generation process,
while \cref{tab:support} isolates support scoring.
Computing exact responses on every frame increases the reported
total to 161.3 minutes.
Under the assumed hardware scaling, the estimated costs of EM
and TAP are 156.1 and 315.8 minutes, respectively.
Random noise, AR and SynPer have reported or estimated costs
below 9 minutes.
The cross-GPU estimates are uncalibrated and do not establish
a hardware-controlled runtime ranking.
The data owner incurs this cost once before release.

\begin{table}[t]
  \centering
  \scriptsize
  \caption{\textbf{Computational costs on A6000.}
  Costs for protecting 52,970 LIBERO-Spatial frames.
  \method includes pattern construction, support scoring
  and release writing, with the cost of exact responses on
  10\% of episodes approximated from the full-suite runtime.}
  \label{tab:cost}
  \setlength{\tabcolsep}{4pt}
  \renewcommand{\arraystretch}{1.12}
  \begin{tabular}{lcc}
    \toprule[1.1pt]
    \rowcolor{tabhead}
    \textbf{Release} 
    & \textbf{Minutes} & \textbf{ms per frame} \\
    \midrule
    Random noise & 8.4 & 9.5 \\
    \rowcolor{tabzebra}
    EM~\cite{huang2021unlearnable}
    & 156.1 & 176.9 \\
    TAP~\cite{fowl2021adversarial}
    & 315.8 & 357.7 \\
    \rowcolor{tabzebra}
    AR~\cite{sandoval2022autoregressive}
    & 6.0 & 6.8 \\
    SynPer~\cite{yu2022availability}
    & 7.2 & 8.2 \\
    \midrule
    \method with exact responses on every frame
    & 161.3 & 182.7 \\
    \rowcolor{tabours}
    \textbf{\method}
    & \textbf{68.9} & \textbf{78.0} \\
    \bottomrule[1.1pt]
  \end{tabular}
\end{table}

\section{Ethics Statement}
\label{sec:ethics}

The intended use is to help data owners share demonstrations for evaluation while discouraging unauthorized training.
The method leaves demonstrated actions and instructions unchanged and does not optimize for a selected robot behavior.

%% file: tables/real.tex
\setlength{\tabcolsep}{4pt}
\renewcommand{\arraystretch}{1.12}
\begin{tabular}{lcccc}
\toprule[1.1pt]
\rowcolor{tabhead} \textbf{Release} & \textbf{Held-out error} & \textbf{Ratio} & \textbf{$\Delta E$} & \textbf{Drift} \\
\midrule
Clean & 0.132 & 1.00 & -- & -- \\
\rowcolor{tabzebra} Random noise & 0.130 & 0.99 & \textminus{}1.1 & 1.1 \\
EM~\cite{huang2021unlearnable} & 0.250 & 1.90 & \textminus{}1.0 & 1.8 \\
\rowcolor{tabzebra} TAP~\cite{fowl2021adversarial} & 0.128 & 0.98 & \textminus{}2.2 & 1.9 \\
AR~\cite{sandoval2022autoregressive} & 0.128 & 0.97 & \textminus{}0.1 & 0.1 \\
\rowcolor{tabzebra} SynPer~\cite{yu2022availability} & 0.126 & 0.96 & \textminus{}0.3 & 0.1 \\
\rowcolor{tabours} \textbf{\method} & \textbf{0.300} & \textbf{2.28} & \textbf{0.0} & \textbf{0.0} \\
\bottomrule[1.1pt]
\end{tabular}

%% file: tables/transfer.tex
\setlength{\tabcolsep}{5pt}
\renewcommand{\arraystretch}{1.12}
\begin{tabular}{lcc}
\toprule[1.1pt]
\rowcolor{tabhead} \textbf{Trained policy} & \textbf{Clean} & \textbf{\method} \\
\midrule
SmolVLA & 74.2 & 7.0 \\
\rowcolor{tabzebra} OpenVLA & 73.0 & 26.0 \\
\bottomrule[1.1pt]
\end{tabular}

%% file: tables/ablation.tex
\setlength{\tabcolsep}{5pt}
\renewcommand{\arraystretch}{1.12}
\begin{tabular}{lccc}
\toprule[1.1pt]
\rowcolor{tabhead} \textbf{Variant} & \textbf{SR} & \textbf{$\Delta E$} & \textbf{Drift} \\
\midrule
Random tile patterns & 27.0 & 0.7 & 1.9 \\
\rowcolor{tabzebra} Per-patch color offsets & 12.8 & 0.0 & 2.2 \\
Uniform support ($\rho=1$) & 0.4 & 15.1 & 17.9 \\
\rowcolor{tabours} \textbf{\method} & \textbf{7.0} & \textbf{2.0} & \textbf{5.4} \\
\bottomrule[1.1pt]
\end{tabular}

%% file: tables/robust.tex
\setlength{\tabcolsep}{5pt}
\renewcommand{\arraystretch}{1.12}
\begin{tabular}{lcc}
\toprule[1.1pt]
\rowcolor{tabhead} \textbf{Operation} & \textbf{\method} & \textbf{Luminance variant} \\
\midrule
None & 7.0 & 0.2 \\
\rowcolor{tabzebra} Training-time augmentation & 10.2 & -- \\
Gaussian blur ($\sigma=1$) & 12.2 & -- \\
\rowcolor{tabzebra} 3-bit depth & 30.4 & -- \\
Grayscale & 23.4 & 0.0 \\
\bottomrule[1.1pt]
\end{tabular}

%% file: tables/hyper.tex
\setlength{\tabcolsep}{5pt}
\renewcommand{\arraystretch}{1.12}
\begin{tabular}{llccc}
\toprule[1.1pt]
\rowcolor{tabhead} \textbf{Parameter} & \textbf{Value} & \textbf{SR} & \textbf{$\Delta E$} & \textbf{Drift} \\
\midrule
$K$ & 4 & 32.2 & 0.8 & 4.0 \\
 & 8 & 3.8 & 1.1 & 4.6 \\
\rowcolor{tabours}  & 17 & 7.0 & 2.0 & 5.4 \\
\arrayrulecolor{black!45}\hline\arrayrulecolor{black}
$\rho$ & 0.25 & 5.6 & 0.4 & 3.0 \\
\rowcolor{tabours}  & 0.5 & 7.0 & 2.0 & 5.4 \\
 & 1 & 0.4 & 15.1 & 17.9 \\
\arrayrulecolor{black!45}\hline\arrayrulecolor{black}
$\epsilon$ & 2/255 & 10.2 & 0.2 & 1.6 \\
\rowcolor{tabours}  & 4/255 & 7.0 & 2.0 & 5.4 \\
 & 8/255 & 0.2 & 11.9 & 18.5 \\
\bottomrule[1.1pt]
\end{tabular}

%% file: tables/share.tex
\setlength{\tabcolsep}{4pt}
\renewcommand{\arraystretch}{1.12}
\begin{tabular}{lccccc}
\toprule[1.1pt]
\rowcolor{tabhead} \textbf{Share} & \textbf{Overlap} & \textbf{Admitted} & \textbf{$\Delta E$} & \textbf{Drift} & \textbf{SR} \\
\midrule
1\% & 77.9 & 0.20 & 2.3 & 5.4 & 1.4 \\
\rowcolor{tabzebra} 5\% & 80.6 & 0.17 & 2.2 & 5.2 & -- \\
25\% & 83.3 & 0.13 & 1.2 & 4.6 & -- \\
\rowcolor{tabours} \textbf{10\%} & \textbf{81.5} & \textbf{0.16} & \textbf{1.9} & \textbf{5.0} & \textbf{7.0} \\
\bottomrule[1.1pt]
\end{tabular}

%% file: tables/support.tex
\setlength{\tabcolsep}{3pt}
\renewcommand{\arraystretch}{1.12}
\begin{tabular}{lcccc}
\toprule[1.1pt]
\rowcolor{tabhead} \textbf{} & \textbf{Mean map} & \textbf{Other suites} & \textbf{10\% episodes} & \textbf{Exact} \\
\midrule
Overlap (\%) & 69.3 & 75.2 & 81.5 & 100 \\
\rowcolor{tabzebra} Admitted response & 0.37 & 0.26 & 0.16 & 0.07 \\
Error change (\%) & -- & +2.8 & +2.0 & +1.3 \\
\rowcolor{tabzebra} Drift (\%) & -- & 6.3 & 5.4 & 4.6 \\
Success (\%) & -- & -- & 7.0 & 8.8 \\
\rowcolor{tabzebra} Time (min) & 0 & 2 & 42 & 137 \\
\bottomrule[1.1pt]
\end{tabular}

%% file: tables/dynamics.tex
\setlength{\tabcolsep}{4pt}
\renewcommand{\arraystretch}{1.12}
\begin{tabular}{llccc}
\toprule[1.1pt]
\rowcolor{tabhead} \textbf{Trained on} & \textbf{Steps} & \textbf{Clean frames} & \textbf{Protected frames} & \textbf{Drift} \\
\midrule
Clean release & 5k & 0.320 & 0.320 & 0.0002 \\
\rowcolor{tabzebra}  & 10k & 0.064 & 0.064 & 0.0003 \\
 & 15k & 0.027 & 0.027 & 0.0002 \\
\rowcolor{tabzebra}  & 20k & 0.024 & 0.024 & 0.0003 \\
\method release & 10k & 0.785 & 0.031 & 0.7541 \\
\rowcolor{tabzebra}  & 20k & 0.742 & 0.013 & 0.7270 \\
\bottomrule[1.1pt]
\end{tabular}